\documentclass[10pt,twocolumn,letterpaper]{article}

\usepackage{cvpr}
\usepackage{times}
\usepackage{epsfig}
\usepackage{graphicx}
\usepackage{amsmath}
\usepackage{amssymb}

\usepackage{epstopdf}

\usepackage{bbm}
\usepackage{dsfont}
\usepackage{float}
\usepackage{algorithmic}
\usepackage{algorithm}
\usepackage{subfigure}
\usepackage{color,xcolor}

\usepackage{multirow}
\usepackage[keeplastbox]{flushend}

\graphicspath{{./figs/}}

\DeclareMathOperator*{\argmax}{argmax}                                          
\DeclareMathOperator*{\argmin}{argmin}         
           
\newcommand{\comment}[1]{}

\newcommand*{\affmark}[1][*]{\textsuperscript{#1}}

\usepackage[pagebackref=true,breaklinks=true,letterpaper=true,colorlinks,bookmarks=false]{hyperref}

\cvprfinalcopy 

\def\cvprPaperID{2606} 

\ifcvprfinal\fi
\begin{document}

\title{Model-based Iterative Restoration for Binary Document Image Compression\\
 with Dictionary Learning}

\author{Yandong Guo\affmark[1]
\thanks{This research work was done 
when Yandong Guo 
and Cheng Lu were Ph.D. students at Purdue University.
This paper is published at CVPR 2017.}\hfill
Cheng Lu\affmark[2] \hfill
Jan P. Allebach\affmark[3]
\hfill 
Charles A. Bouman\affmark[3]\\
\affmark[1]Microsoft Research\hfill
\affmark[2]Sony Electronics Inc. \hfill 
\affmark[3]Purdue University at West Lafayette\hfill \\
{\tt\small yandong.guo@microsoft.com, cheng.lu@am.sony.com, \{allebach, bouman\}@purdue.edu}
}

\maketitle

\begin{abstract}
The inherent noise in the observed (e.g., scanned) binary document image 
degrades the image quality and harms the compression ratio
through breaking 
the pattern repentance
and adding entropy to the document images. 
In this paper, 
we design a cost function in Bayesian framework with dictionary learning. 
Minimizing our cost function produces a restored image which
has better quality than that of the observed noisy image, 
and
a dictionary for representing and encoding the image. 
After the restoration, 
we use this dictionary (from the same cost function) 
to \textbf{encode the restored image} 
following the symbol-dictionary framework 
by JBIG2 standard 
\textbf{with the lossless mode}. 
Experimental results with a variety of document images 
demonstrate that our method improves the image quality compared with the observed image, 
and simultaneously improves the compression ratio. 
For the test images with synthetic noise, 
our method reduces the number of flipped pixels by $48.2\%$
and improves the compression ratio by $36.36\%$ as compared with the best encoding methods. 
For the test images with real noise, 
our method 
visually improves the image quality, 
and outperforms the cutting-edge method 
by $28.27\%$ in terms of the compression ratio.
\end{abstract}

\section{Introduction}
To have binary document images with better quality and smaller sizes are the two goals that have been pursued for decades. 
The high compression ratio of document images mainly relies on
the information redundancy embedded in the repeated patterns of the document image, 
as well as an intelligent way to leverage this pattern repentance.

\begin{figure}[!t]
\begin{center}
\hspace*{\fill}
\includegraphics[width=0.99\linewidth]{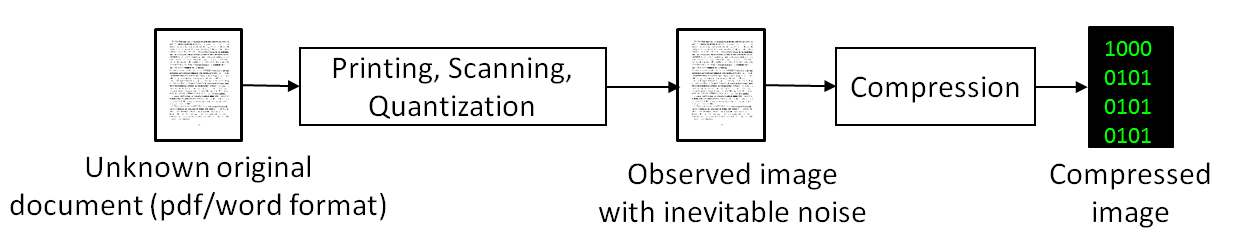} \\
\hspace{0.2cm}
\subfigure[Unknown, original image]{\includegraphics[width=0.235\linewidth]{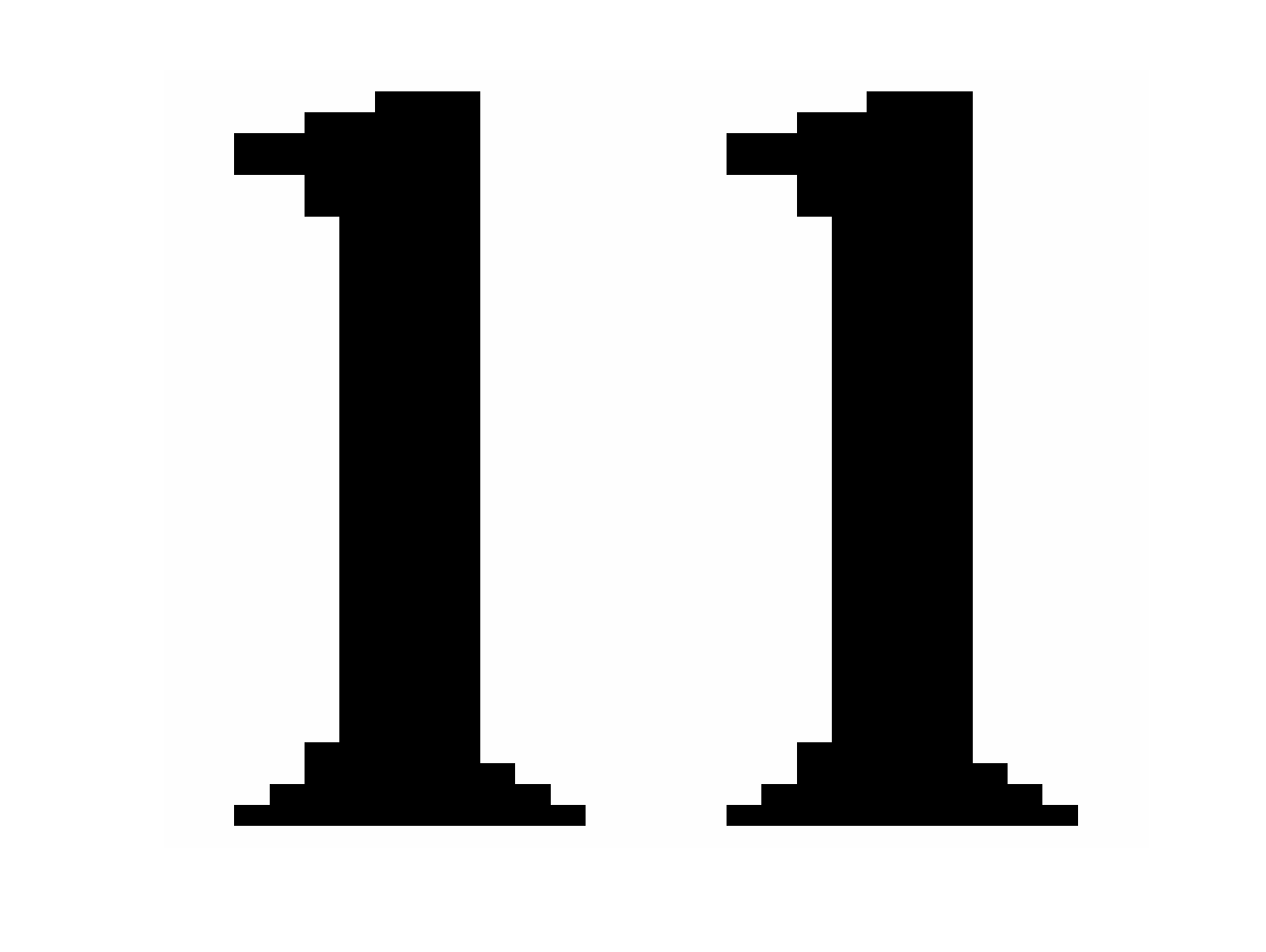}}
\hspace{1.5cm}
\subfigure[Input: noisy observation]{\includegraphics[width=0.235\linewidth]{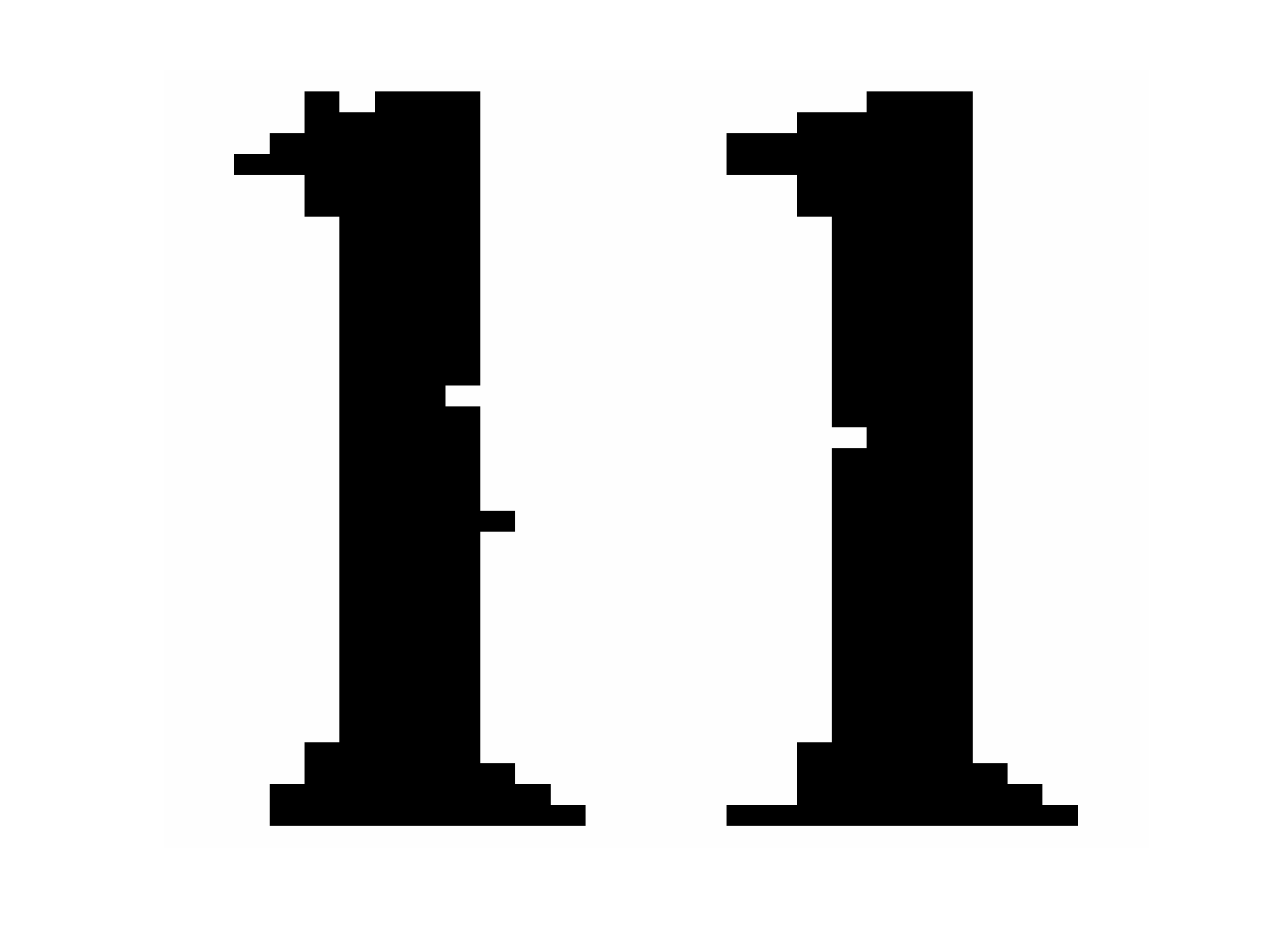}}
\hspace{0.3cm}
\subfigure[Restored by our method]{\includegraphics[width=0.235\linewidth]{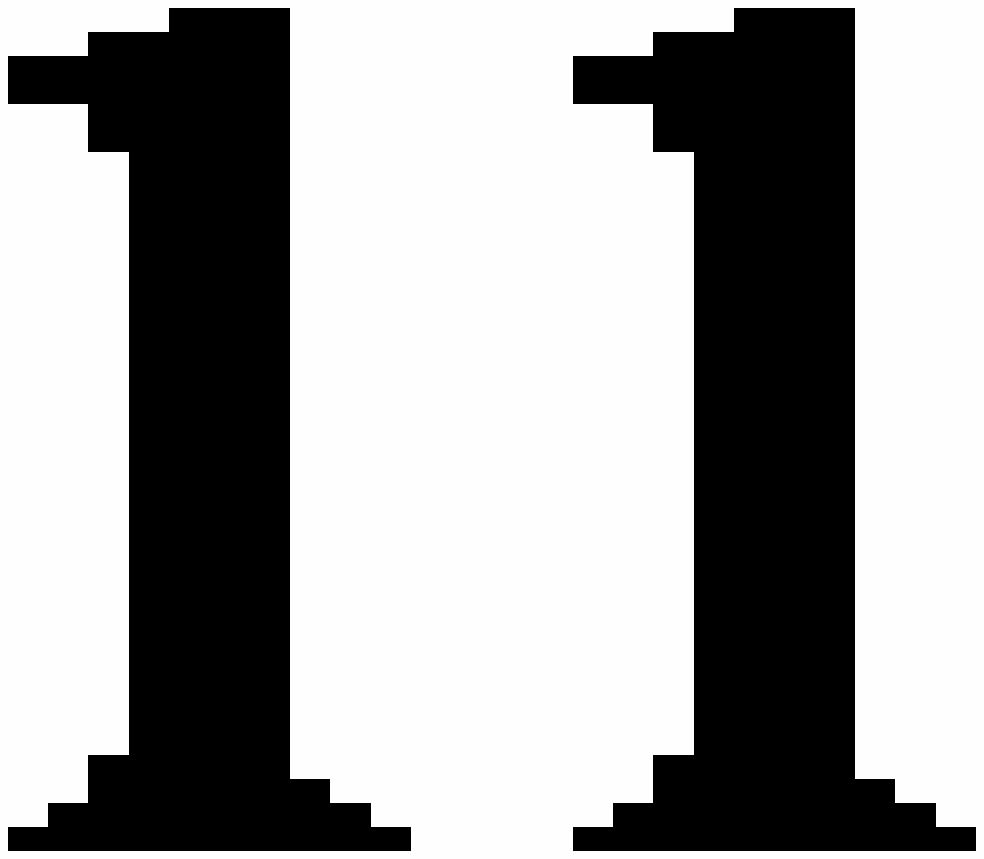}}
\hspace*{\fill}
\end{center}
\caption{
The imaging and compression pipeline for document images. 
In the bottom area of the figure, 
we zoom in the document image to visualize
the details of the two characters ``l''.
As shown in subfigure (b),
the input of our system
contains noise
inevitably
introduced 
by using the imaging device (scanners, cameras, etc.). 
Our method restores the input noisy image and compresses the restored image. 
As shown in subfigure (c), 
our method successfully removes the noise
and
maintains/recovers the very fine details (one-pixel width stroke). 
Moreover, we present 
in Sec. \ref{sec:exp} that
encoding 
our restored images, 
compared with encoding the observed images,
improves the compression ratio
by
$36.36\%$
in the synthetic noise setup
and $28.27\%$ in the real noise setup. 
}
\label{fig:flowchart} 
\end{figure}

Unfortunately, 
when the document image is obtained through scanning or other imaging devices, 
noise is inevitably introduced. 
This inherent noise breaks the pattern repentance, 
increases the entropy, 
and therefore lowers the compression ratio. 
As shown in Figure \ref{fig:flowchart} (b), 
the same letter ``l'' look different from each other in the observed image,
though they used to have the same typeface and font size
in the original image in Figure \ref{fig:flowchart} (a).
The stroke lost its smooth boundary in the observed image. 
In most of the scenarios, only the observed noisy images 
are available. More examples are shown in Figure \ref{fig:recon} and \ref{fig:realQuality}.  

Conventionally, there are two options to compress these observed images. 
In the first option, people encode the observed image as it is (lossless mode). 
In this case, the quality of the compressed image is equal to the quality of the observed image,
while a significant portion of the bits are used to store nothing but noise. 
Option two is the lossy mode, which tends to have high compression ratio, 
but would typically make the image quality worse
than the quality of the observed image input, 
or even introduces semantic errors to the document image.

In this paper, 
we solve the above problem from a different perspective. 
We propose a restoration method to improve the quality of the observed noisy image, 
and simultaneously favors the compression ratio (compared with directly encoding the observed image). 
The intuition is that the pattern repentance of the observed noisy document image is naturally recovered during our image restoration procedure, 
and this pattern repentance benefits the compression ratio. 
Our method is summarized in two steps. 
First, we restore the image by minimizing a cost function (Eq. \ref{eq:MAP}) in
Bayesian framework.
Second, after the restoration, we use the same dictionary for restoration to encode the restored document image.

Our cost function is the summation of a likelihood term
and a prior term,  
\begin{equation}
\label{eq:MAP}
\{\hat{\bf x}, \hat{\bf D} \}= \argmin_{\bf x, \bf D} \{ -\log p({\bf y}|{\bf x}) -\log p({\bf x} | {\bf D} ) -\log p( {\bf D} )\} \, .
\end{equation}
The likelihood term $-\log p({\bf y}|{\bf x})$ is used to simulate a typical imaging pipeline (from the unknown, noise-free image ${\bf x}$ 
to the observed noisy image ${\bf y}$), 
while
the prior term (the rest of the cost) is designed 
to encourage the image ${\bf x}$ to be sparsely represented by a dictionary ${\bf D}$ of limited size. 
We learn this dictionary globally from the observed noise image, 
and leverage 
the non-local information embedded in the dictionary to improve the image quality 
and recover the repeated patterns of the document image. 

More specifically, 
we learn our dictionary 
in the conditional entropy estimation (CEE) space in \cite{Yandong:JBIG2CEE}, 
and leverage CEE to calculate the sparse representation cost $-\log p({\bf x} | {\bf D} )$ in the prior term. 
The previous art \cite{Yandong:JBIG2CEE}
demonstrates that
the distribution of binary signals is better modeled in the CEE space (compared with that in the Euclidean space), 
and the CEE space has significant advantages
in evaluating the amount of the information contained in image patches given the associated dictionary entries. 

After the restoration, we first encode the dictionary
$\hat{{\bf D}}$ estimated in the cost function in Eq. (\ref{eq:MAP}), 
and then encode the restored image $\hat{\bf x}$ using this dictionary as a reference.
Our encoding follows the JBIG2 lossless encoding standard \cite{st:kx}. 

Since our sparse representation cost $-\log p({\bf x} | {\bf D} )$ is calculated 
by estimating the information entropy in the image given the dictionary, 
and we use the same dictionary for restoration and compression, 
our prior term in Eq. (\ref{eq:MAP}) has the capability of approximating the number of bits required to encode the image. 
Therefore, minimizing the cost function in the preprocessing step 
does not only improve the image quality, 
but also numerically reduces the approximated file size required to encode the image, with the constraint $-\log p({\bf y}|{\bf x})$.
To the best of our knowledge, this is the first time that the same dictionary is shared by restoration and compression.


We conduct experiments with test images with synthetic noise and real noise. 
Experimental results 
demonstrate that 
our restored image has higher quality than that of the observed image, 
and encoding the restored image generates higher compression ratio compared with directly encoding the input observed image. 

The contribution of our paper is summarized as follows. 
\begin{itemize}
\item
We design a cost function in Eq. (\ref{eq:MAP}). 
This cost function
is used to model image restoration, 
and also approximate the number of bits required to encode the image. 
Minimizing this cost function simultaneously improves the quality 
of the observed (e.g., scanned) document image, 
and improves the compression ratio. 




\item 
We learn our dictionary in the conditional entropy space, where the binary signal distribution is better modeled \cite{Yandong:JBIG2CEE}. 
\item 
To the best of our knowledge, it is the first time that the same dictionary is used for restoration and compression. 

\item 
Our bistream is compliant with the JBIG2 standard. 
\end{itemize}

The paper is organized as follows. 
In Sec. \ref{sec:related}, we review some of the most related work. 
In Sec. \ref{sec:model}, we describe our mathematical model for both imaging and prior learning. 
In Sec. \ref{sec:opt}, the method to optimize our model is presented. 
Experimental results for the test images with synthetic and real noise are shown in Sec. \ref{sec:exp}. 

\section{Related works}
\label{sec:related}


Since we have not yet seen much effort published in 
optimizing restoration quality and compression ratio together,  
we review compression and restoration methods separately. 

\subsection{JBIG2 encoding}

After we finish preprocessing the image with Eq. \ref{eq:MAP}, 
we 
encode the {\bf restored image}
with the symbol-dictionary framework defined in the JBIG2 compression standard with the {\bf lossless mode}, 
developed by the Joint Bi-level Image Experts Group \cite{st:kx}. 
The JBIG2 compression standard produces higher compression ratios than the previous standards, 
such as T.4, T.6, and T.82 \cite{st:kxg3, st:kxg4, st:kxjbig1, HR:standards, TK:com}, 
through the symbol-dictionary framework. 
A typical JBIG2 encoder
works by first separating the document images into repeated connected components, called symbols. 
Then, the encoder encodes the learned dictionary entries as part of the bitstream, 
then encode the image using the dictionary entries as reference
\cite{Fum, Howard:1998vn, JBIG2:Arps, P:Lossless, ye:dictionary}. 

With the lossless mode, 
all the difference between the image patch and the 
associated
dictionary entry is entropy encoded. 
The conventional JBIG2 lossless encoders compress the observed noisy image. 
In this case, 
the inherent noise tends to increase entropy in the image and consumes extra bits when the document image is encoded. 
On the contrary, our method compress the restored image to produce better quality and higher compression ratio. 

While all the conventional JBIG2 encoders compress the observed image, 
some encoders achieve higher compression by 
better dictionary learning. 
The dictionary learning typically consists of two critical tasks; 
one is to construct the dictionary, the other one is to select the best dictionary entry for a given image patch (symbol). 
These two tasks could be done alternatively or simultaneously. 

Typically, the dictionary entry selection for a given symbol is accomplished
by minimizing a measure of dissimilarity between the symbol and the dictionary entry.  
Dissimilarity measures widely used in JBIG2 include 
the Hamming distance, known as XOR \cite{holt:a}, and weighted Hamming distance, known as WXOR \cite{P:Combined, M:JBIG2}. 
The weighted Hamming distance is calculated
as the weighted summation of the difference between a symbol bitmap and a dictionary entry bitmap. 
Zhang, Danskin, and Yong have also proposed a dissimilarly measure
based on cross-entropy which is implemented as WXOR with specific weights \cite{Zhang:E, Zhang:E2}. 
The XOR has the lowest computational cost, while WXOR and cross-entropy methods are more widely 
used 
because they are more sensitive to clustered errors and can achieve lower substitution error \cite{M:JBIG2, F:JBIG2T}.
These days, to evaluate the dissimilarity between the symbol and the dictionary entry using conditional probability estimation  
shows great potential in \cite{information:theoretic, Yandong:JBIG2CEE, Yandong:JBIG2Multi}. 
The OCR-based method needs extensive training, and is sensitive to font and/or language type, so is beyond the discussion in this paper. 

For dictionary construction,    
various methods have been proposed.
These methods typically cluster the symbols into groups, 
according to a dissimilarity measure, 
using K-means clustering or a minimum spanning tree \cite{ye:dictionary, ye:symbol, ye:fast}.
Within each group, one dictionary entry is used to represent all the symbols of that group. 

Note that 
the JBIG2 standard also provides a lossy option. 
Different from the typical definition of ``lossy'' in JPEG or typical video coding,
the lossy-JBIG2 refers to replacing the image symbols 
with their associated dictionary entries. 
The lossy option is very risky to use due to the following two types of potential quality degradation. 
The first one is called substitution error, 
which happens when the symbol is replaced by a dictionary entry with different semantic meaning. 
For example, the  letter ``c'' could easily be replaced by the letter ``o'', especially in the low resolution scanning condition. 
Though many methods, including \cite{P:Combined, M:JBIG2}, have been proposed to control the substitution error, 
we have yet to see any of them claims zero error rate. 
The second type of quality degradation happens  
when the symbol is substituted by a dictionary entry with the same semantic meaning, but lower quality. 
However, there has 
not been much effort in this field to ensure the dictionary entry has better quality than the symbols to be replaced. 
Due to these reasons, we do not consider the lossy mode of JBIG2 encoder in this paper. 
\subsection{Image restoration}

The paper \cite{Peyman:MIP} provides a very comprehensive review from the perspective of filtering. 
Among all these methods, 
model-based reconstruction/restoration methods
with a Markov random field (MRF) prior
\cite{Eri:MRF, MM:restoration, tip_Unified}, 
offers very robust results. 
Moreover, 
recent methods utilizing non-local information obtain the cutting-edge performance in restoring gray/color images, 
\eg, \cite{Zoran11, Yu12, nonLocalMean, nonLocalMeanRestoration, BM3D:Restoration, KSVD:denoising, KSVD:denoising2, KSVD:Restoration}, 
and promising results in various reconstruction applications, \eg, \cite{Wang12, Zhang13, Xu12, Jin12}. 

Extra work is needed to transfer these methods designed for gray image restoration to our problem. 
One major reason is that the distortion in binary document images has different patterns which can not be well approximated by Gaussian distribution (the implicit assumption in most of the restoration works above).  
The non-local information of the binary document image need
to be used in a better way. 
Moreover, none of these above restoration methods are designed to improve the compression ratio. 
We solve these problems in this paper by 
optimizing one cost function, 
which simultaneously takes care of image quality and compression ratio. 


\section{Statistical model}
\label{sec:model}
Let ${\bf x} \in \{0, 1\}^{K}$ denote the unknown noise-free image,
the vector ${\bf y} \in \{0, 1\}^{K}$ denote the observed image, 
we obtain the restored image to be encoded 
by minimizing the cost function in Eq. \ref{eq:MAP}. 
Details of each term in Eq. \ref{eq:MAP} are presented in the following subsections. 

\subsection{Forward model for the likelihood term}
Given the distortion-free unknown image ${\bf x} \in \{0, 1\}^{K}$,
the observed image ${\bf y} \in \{0, 1\}^{K}$ has the following likelihood distribution, 
\begin{equation}
\label{eq:likelihood}
p({\bf y}|{\bf x}) = \prod_k p(y_k|{\bf x}) \, ,
\end{equation}
where, 
 \begin{align}
\label{eq:likelihood_indi}
p(y_k|{\bf x}) &= 1 - | y_k -\mu_k |  \\
\label{eq:likelihood_lowp}
\boldsymbol \mu &= A {\bf x} \, .
\end{align}
The term $| y_k -\mu_k |$ is the absolute value of $y_k -\mu_k $. 
In the above equations, 
Eq. (\ref{eq:likelihood_lowp})
is based on the low pass assumption of printing and scanning
due to the limited resolution of these procedures. 
We formulate this low pass filter using the matrix $A \in \Re^{K \times K}$, 
each row of which performs a low pass filter to the image ${\bf x}$, 
and denote the intermediate image to be $\boldsymbol \mu \in [0, 1]^{K}$. 
We constrain the matrix $A$ to be sparse to achieve low computational cost, 
and also constrain $A$ to be circulant to achieve homogeneous filtering to the image ${\bf x}$.  
Moreover, 
we propose the following constraint on each row of $A$
to ensure there is no energy change introduced by filtering. 
\begin{equation}
\label{eq:Acons}
\sum_l A_{k, l} = 1 \, .
\end{equation}


Equation (\ref{eq:likelihood_indi}) describes the conditional probability distribution of the $k^{th}$ pixel $y_k$. 
Since the pixel $y_k$ has the value of either $1$ or $0$, 
we can express Eq. (\ref{eq:likelihood_indi}) as follows, 
\begin{align}
\label{eq:miu1}
p(y_k = 1| \mu_k) &= \mu_k  \\
\label{eq:miu0}
p(y_k = 0| \mu_k) &= 1- \mu_k \,.
\end{align}
The above Eq. (\ref{eq:miu1}) and (\ref{eq:miu0}) show that Eq. (\ref{eq:likelihood_indi}) is a valid probability distribution.  
Moreover, Eq. (\ref{eq:miu1}) and (\ref{eq:miu0}) demonstrate our intuitions to design the likelihood function:
if the pixel $\mu_k$ in the intermediate image has a large value closer to $1$, we have larger chance to obtain $y_k = 1$; 
while if the pixel $\mu_k$ has a small value closer to $0$, we have larger chance to obtain $y_k = 0$. 

With the two models for low pass filtering in Eq. (\ref{eq:likelihood_lowp}) and following quantization described in Eq. (\ref{eq:miu1}) and (\ref{eq:miu0}),
we establish the likelihood function in Eq. (\ref{eq:likelihood}) based on the assumption that 
each of the pixels in the observed image ${\bf y}$ are 
conditionally independent distributed, given the latent image ${\bf x}$. 
\begin{align}
\label{eq:likelihood_express}
p({\bf y}|{\bf x}) =  \prod_k \left(1 - | y_k - \sum_l A_{k,l} {x_l} | \right)
\end{align}

Here, for both simplicity reason and the model generality, we assume that the probability distribution of the pixel $y_k$ is only determined
by the pixel value of $\mu_k$. 
For a specific quantization algorithm, such as error diffusion, we can update the likelihood function accordingly. 

\subsection{Prior model with dictionary learning}

We design the prior term in Eq. (\ref{eq:MAP}) as follows, 
\begin{align}
\label{eq:prior}
-\log p({\bf x}| {\bf D}) - p({\bf D})  \propto &-\sum_i \log p(B_i {\bf x}| {\bf d}_{f(i)}; \boldsymbol \phi)  \nonumber \\
&- \sum_j \log p({\bf d}_j) \, . 
\end{align}

In the first summation term, 
the term $p(B_i {\bf x}| {\bf d}_{f(i)}; \boldsymbol \phi)$ is the conditional probability of the $i^{th}$ symbol
given the $f(i)^{th}$ dictionary entry ${\bf d}_{f(i)} \in {\bf D}$, parameterized by $\boldsymbol \phi$. 
The matrix $B_i$ is the operator used to extract the $i^{th}$ patch (called the $i^{th}$ \textit{symbol}) in the image, 
and $j=f(i)$ denote the function that maps each individual symbol, $B_i{\bf x}$, 
to its corresponding dictionary entry, ${\bf d}_j \in {\bf D}$.
For notation simplicity, 
we define
\begin{equation}
{\bf s}_i = B_i {\bf x} \, .
\end{equation}
The second summation term is the penalizer of the dictionary size. 

Our prior design has two meanings. 
One is for restoration: to 
encourage the image to be represented by a dictionary with limited size. 
The other one is to approximate the number of the bits required to encode the image. 

More specifically, 
the variable $\boldsymbol \phi$ is introduced to parameterize the conditional probability $p({\bf s}_i| {\bf d}_{j};  \boldsymbol \phi)$. 
We do 
not calculate Euclidean distance between the image batch and the associated dictionary entry
as the log of the conditional probability
because the distortion in document binary images typically does not follow the independently identically Gaussian distributed assumption well
(which is the prerequisite of using Euclidean distance).
Intuitively speaking, 
the benefit of using  $\boldsymbol \phi$ to parameterize the conditional probability is that we can have larger weight for the rare distortion patterns, 
while have smaller weight for the common distortion patterns, through a rigid optimization procedure over $\boldsymbol \phi$. 
Different weights for different distortion patterns
introduce a good approximation to the amount of information needed to be encoded for the symbol given the associated dictionary entry
\cite{Yandong:JBIG2CEE, Yandong:JBIG2Multi}. 
This good approximation benefits the dictionary entry selection and construction, which eventually benefits the restoration and the compression. 
More detailed experimental results in Sec. \ref{sec:exp} further demonstrate advantages in estimating $\boldsymbol \phi$
in aspects of both compression and restoration. 

\begin{figure}[!t]
\begin{center}
\hspace*{\fill}
\subfigure[Neighbors in symbol]{\includegraphics[width=0.46\linewidth]{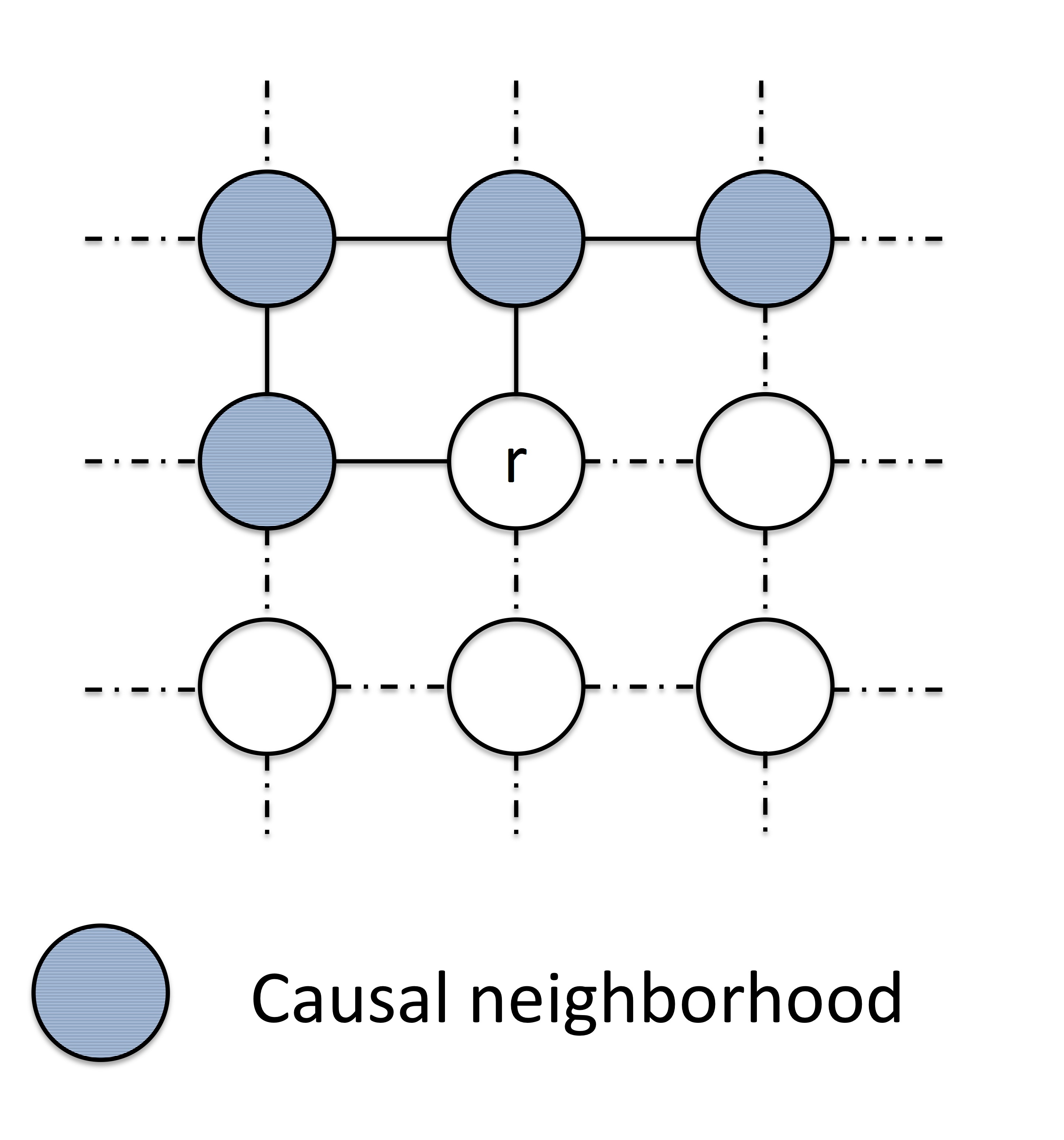}}
\hspace*{\fill}
\subfigure[Neighbors in dictionary entry]{\includegraphics[width=0.46\linewidth]{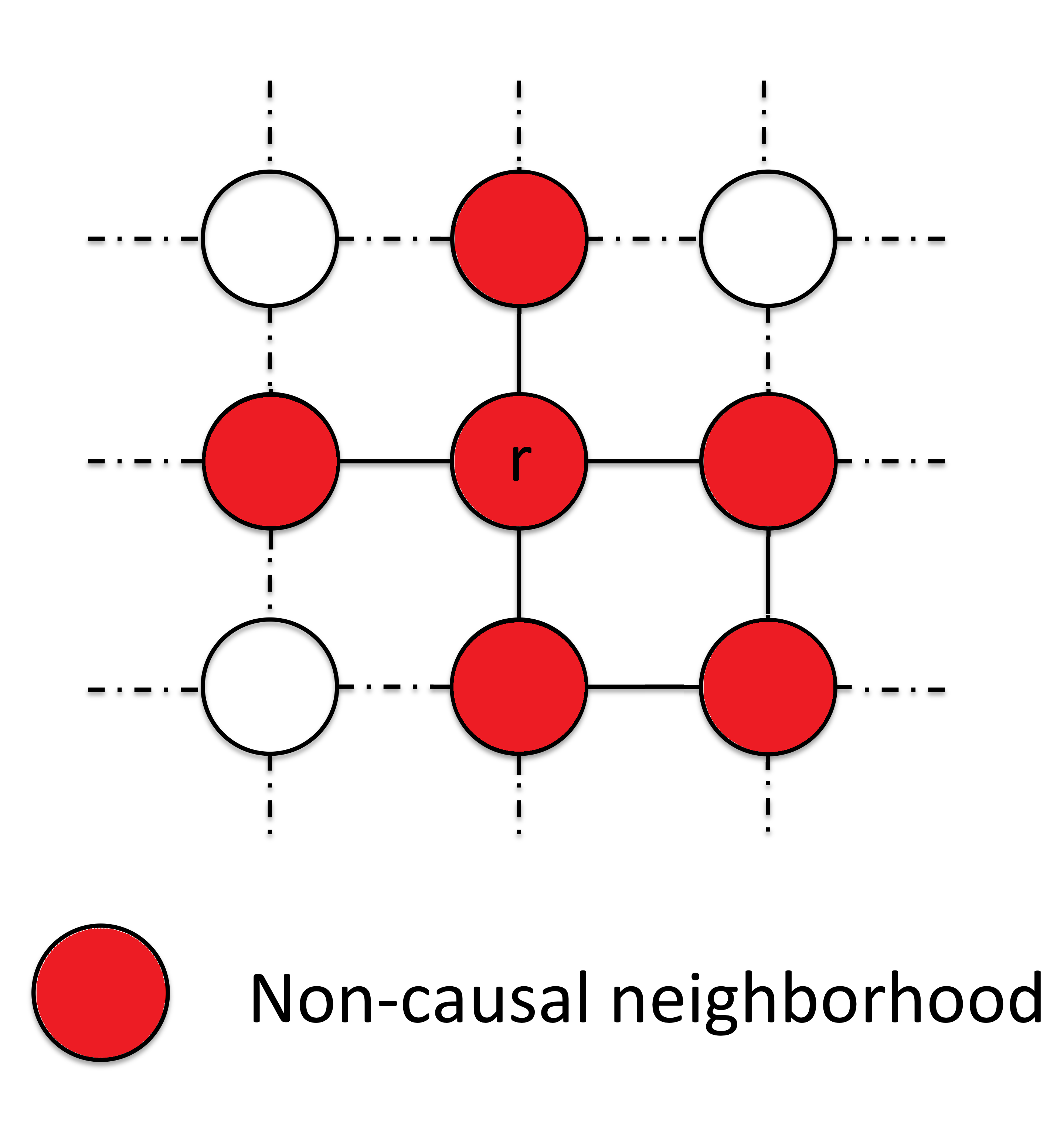}}
\hspace*{\fill}
\end{center}
\caption{
The $4$ causal neighborhood pixels of $s_{i}(r)$ in ${\bf s}_i$, 
and the $6$ non-causal neighborhood pixels of $d_{j}(r)$ in ${\bf d}_{j}$. 
Note that this is not the only neighborhood system we can use. 
We choose the neighborhood system which is also used in the JBIG2 standard \cite{st:kx}, 
but estimate the conditional probability in a different way, as described in Sec. \ref{sec:opt}.   
}
\label{fig:neighborhood} 
\end{figure}

We briefly review how we model the conditional probability $p({\bf s}_i| {\bf d}_{j}; \boldsymbol \phi) $. 
The conditional probability $p({\bf s}_i| {\bf d}_{j}; \boldsymbol \phi) $
can have a very complicated form, 
since both ${\bf s}_{i}$ and ${\bf d}_{j}$
are high dimensional random variables. 
This makes 
the parameter vector $ \boldsymbol \phi$ contain too many elements to be estimated.
To solve this problem, 
we model $p({\bf s}_i| {\bf d}_{j};  \boldsymbol \phi) $
as the product of a sequence of simple probability density functions, 
\begin{align}
\label{eq:HPD}
   &p({\bf s}_i| {\bf d}_{j};  \boldsymbol \phi) 
= \prod_{s}
p \left( s_{i}(r)| {\bf c} ({\bf s}_i, {\bf d}_j, r) ;  \boldsymbol \phi \right) \, ,
\end{align}
where 
the term $p \left( s_{i}(r)| {\bf c} ({\bf s}_i, {\bf d}_j, r) ;  \boldsymbol \phi \right)$
is the conditional probability for the $r^{th}$ symbol pixel $s_{i}(r)$
conditioned on its reference context ${\bf c} ({\bf s}_i, {\bf d}_j, r) $,
of which the definition 
is shown in Fig. \ref{fig:neighborhood}. 

Figure \ref{fig:neighborhood} graphically illustrates one example of the structure of the reference context. 
As shown, the  reference context ${\bf c} ({\bf s}_i, {\bf d}_j, r) $ is a 10-dimensional binary vector, 
consisting of $4$ causal neighborhood pixels of $s_{i}(r)$ in ${\bf s}_i$, 
and $6$ non-causal neighborhood pixels of $d_{j}(r)$ in ${\bf d}_{j}$
. 
The decomposition in (\ref{eq:HPD}) is based on the assumption 
that, 
the symbol pixel $s_{i}(r)$, given its reference context ${\bf c} ({\bf s}_i, {\bf d}_j, r)$,
is conditionally independent of its previous (in raster order) symbol pixels except its $4$ casual neighbors. 
This conditional independency design 
makes our decomposition different
from the existing decomposition/factorization methods 
in inference complicated distributions \cite{MFB, Minka:EP, Yandong:REP}. 

With the decomposition in Eq. (\ref{eq:HPD}),  
we further heuristically assume that
for a given document image, 
the natural parameter $\boldsymbol \phi$
in $p \left( s_{i}(r)| {\bf c} ({\bf s}_i, {\bf d}_j, r) ;  \boldsymbol \phi \right)$
is completely determined by the reference context ${\bf c} ({\bf s}_i, {\bf d}_j, r)$. 
Since the symbol pixels 
are binary, 
we model their conditional distribution given a particular reference context
as a Bernoulli distribution, shown as follows,   
\comment{In our CEE method, }
\begin{align}
\label{eq:Ber}
p(s_i(r) | {\bf c} ({\bf s}_i, {\bf d}_j, r);  \boldsymbol \phi  )  &= \phi_c^{1-s_i(r)} (1-\phi_c)^{s_i(r)}  \, ,  
\end{align}
where
the variable
$\phi_c$ denotes
the natural parameter of the Bernoulli distribution  
and fully determined by the value of the reference context vector  
$c = {\bf c} ({\bf s}_i, {\bf d}_j, r)$. In total, this reference context ${\bf c} ({\bf s}_i, {\bf d}_j, r)$ could possibly have $2^{10}$ different values
with our $10$ bit neighborhood system in Fig. \ref{fig:neighborhood}, 
so there are $2^{10}$ parameters to be estimated. 
\begin{equation}
\boldsymbol \phi = \left[\phi_1, \phi_2, \dots, \phi_{1024} \right]^T 
\end{equation}

\section{Optimization}
\label{sec:opt}
With the likelihood distribution in Eq. (\ref{eq:likelihood}),(\ref{eq:likelihood_indi}), and (\ref{eq:likelihood_lowp}), 
and the prior distribution in Eq. (\ref{eq:prior}), 
we obtain the cost function to be optimized as, 
\begin{align}
\label{eq:cost}
\{ \hat{\bf x}, \hat{\bf D}, \hat f , \hat{\boldsymbol \phi}\} & = \argmin_{{\bf x}, {\bf D}, f, \boldsymbol \phi} 
-\sum_k \log ( 1 - | y_k - \sum_l A_{k,l} {x}_l | ) \nonumber \\
- \sum_i & \log p(B_i {\bf x}| {\bf d}_{f(i)}; \boldsymbol \phi) - \sum_j \log p({\bf d}_j) 
\end{align}

We propose to use an alternating optimization strategy. 
First, we initialize the unknown image $\bf x$ by, 
\begin{equation}
{\bf x} \leftarrow {\bf y} \, .
\end{equation}
Then, we update the dictionary $\bf D$, the mapping $f$, parameter $\boldsymbol \phi$, and the unknown image $\bf x$ alternatively. 
Overall structure of our method is listed in Fig. \ref{fig:pseudocode}, while details are provided in the following subsections. 

\begin{figure}[!t]
\framebox{\begin{minipage}[t]{0.97\columnwidth}
\begin{algorithmic}
\STATE MBIR\_DL\_Encoding$\left( {\bf y}\right) \{$
\centerline{\begin{minipage}[t]{0.924\columnwidth}
\STATE  $ \slash\ast \text{Initialization} \ast \slash$
\begin{flalign}
&
{\hat{\bf x}} \leftarrow {\bf y}  & \nonumber \\
&\{ \hat{\bf D}^{(0)}, \hat f^{(0)} \} \leftarrow \text{XOR-OP  } (\hat {\bf x} ) & \nonumber
\end{flalign}
\REPEAT 
\STATE  
Update $\hat{\boldsymbol \phi}$ using (\ref{eq:PhiObject2})
\STATE  
Update $ \hat{\bf D}, \hat f$ using (\ref{eq:dictionary})
\STATE  
Update ${\hat{\bf x}}$ using (\ref{eq:xupdate})
\UNTIL{
Converge \OR Maximum number of iterations reached
}
\STATE
\STATE
{\textbf{Encode}} $\hat{\bf x}$ using JBIG2 with lossless option
\STATE
\RETURN JBIG2 bitstream
\end{minipage}}
\STATE $\}$
\end{algorithmic}
\end{minipage}}
\caption{Pseudocode of our method called model based iterative restoration for compression with dictionary learning (MBIR-DL-Encoding). 
First, as the initial step, 
we initialize the unknown image ${\bf x}$ with the observed image ${\bf y}$. 
Then, we repeat the parameter estimation, dictionary construction, and image restoration for multiple times until converge.  
After convergence, we encode the restored image $\hat{\bf x}$ using the JBIG2 lossless option. 
}
\label{fig:pseudocode}
\end{figure}

\subsection{Dictionary learning}
At the initial stage, we learn a temporary dictionary $\hat{\bf D}$ and mapping $\hat{f}$ from the current image estimation $\hat{\bf x}^{(0)}$. 
During the dictionary learning, we first estimate the parameter $\boldsymbol \phi$, 
\begin{equation}
\label{eq:phi}
\hat{\boldsymbol \phi} = \argmin_{\boldsymbol \phi} 
- \sum_i \log p(B_i \hat{\bf x}| \hat{\bf d}_{\hat{f}(i)}; \boldsymbol \phi) 
-  \log p_{\phi}(\boldsymbol \phi) \,,
\end{equation}
where the term $p_{\phi}(\boldsymbol \phi)$ is proposed to to stabilize the estimation of $\boldsymbol \phi$. 
In this distribution, 
we assume that
all the elements in $\boldsymbol \phi$ are 
independent and identically distributed, 
following Beta distribution, 
\begin{align}
\label{eq:phiprior}
p_{ \phi} ( \boldsymbol \phi) &=  \prod_c \text{Beta}(\phi_c|a, b) \, , \\
\label{eq:priorfactor}
        \text{Beta}(\phi_c|a, b) &= \frac{\Gamma(a+b)}{\Gamma(a)\Gamma(b)}\phi_c^{a-1}(1-\phi_c)^{b-1} \, . 
\end{align}
We set $a = b = 2$.  

With Eq. (\ref{eq:HPD}) and (\ref{eq:Ber}), 
and the prior (\ref{eq:phiprior}) and (\ref{eq:priorfactor}), 
we update Eq. (\ref{eq:phi}) as the following Eq. (\ref{eq:PhiObject2}), which leads 
to an efficient calculation of $\hat{\boldsymbol \phi}$. 
\begin{align}
\label{eq:PhiObject2}
\hat{\boldsymbol \phi} 
= 
\argmax_{\boldsymbol \phi} & \left\{    \sum_{i=1}^N  \sum_r 
\left[1-\hat s_i(r) \right] \log  \phi_{c (\hat {\bf s}_i, \hat {\bf d}_{\hat f(i)}, r)} \right .  \nonumber \\
&+\sum_{i=1}^N  \sum_r 
{\hat s_i(r)} \log \left(1-\phi_{c (\hat {\bf s}_i, \hat {\bf d}_{\hat f(i)}, r)}\right)  \nonumber \\
&+ \left . \sum_c \log \phi_c (1-\phi_c) \right\}
\end{align}


With the estimation of the conditional probability parameter $\hat{\boldsymbol \phi}$ fixed, 
we construct the dictionary $\hat{\bf D}$ and the mapping $\hat f$ using, 
\begin{align}
\label{eq:dictionary}
\{  \hat{\bf D}, \hat f \} \leftarrow \argmin_{ {\bf D}, f} 
&- \sum_i \log p(B_i  \hat{\bf x}| {\bf d}_{f(i)}; \hat{\boldsymbol \phi})  \nonumber \\
&- \sum_j \log p({\bf d}_j) 
\end{align}
We treat this optimization as a clustering problem in entropy space, 
and use unsupervised greedy agglomerative clustering method to build up the dictionary and mapping.

\subsection{Image restoration}
In section, we present our method to restore the image with the dictionary $\hat{\bf D}$ and the mapping $\hat f$ fixed, 
\begin{align}
\label{eq:x}
\hat{\bf x} \leftarrow \argmin_{{\bf x}} 
&-\sum_k \log ( 1 - | y_k - \sum_l A_{k,l} {x}_l | )  \nonumber \\
&- \sum_i \log p(B_i {\bf x}| \hat{\bf d}_{\hat{f}(i)}; \hat{\boldsymbol \phi})
\end{align}
 
Due to the complexity of Eq. (\ref{eq:x}), 
we design an iterative restoration method. 
At each step, we update only one pixel of the unknown image $\bf x$, and keep the rest pixels the same. 
We use $\tilde {\bf x}^u$ to denote the new image with the $u^{th}$ pixel to be updated. 
The value change of the likelihood term (\ref{eq:x}) is simplified as, 
\begin{align}
\label{eq:likelihood_change}
\Delta_1 = 
-\log 
\frac
{\prod_{\{k|A_{k,u}\neq 0\}}  \left( 1 - \| y_k - \sum_l A_{k,l} \tilde {x}_l^u \| \right)}
{\prod_{\{k|A_{k,u}\neq 0\}} \left( 1 - \| y_k - \sum_l A_{k,l} {x}_l \| \right)}
\end{align}
Note that only the rows in $A$ of which the $u^{th}$ element is nonzero need to be evaluated. 


With the image update, the value change of the prior term is 
\begin{align}
\label{eq:prior_change}
\Delta_2 = - \sum_i \log p(B_i \tilde {\bf x}^u| \hat{\bf d}_{\hat f(i)}; \hat{\boldsymbol \phi}) \nonumber \\
+ \sum_i \log p(B_i {\bf x}| \hat{\bf d}_{\hat f(i)}; \hat{\boldsymbol \phi})  \, , 
\end{align}
which is efficiently calculated because only the symbol which contains the updated pixel $\tilde x_u$ needs to be considered. 
Suppose $s_{i(u)}(r)$ is the $i(u)^{th}$ symbol which contains the updated $u^{th}$ pixel, and 
the changed pixel has a index $r$, 
we can rely on the decomposition in Eq. (\ref{eq:HPD}) 
to simplify Eq. (\ref{eq:prior_change}) as, 
\begin{align}
\label{eq:prior_changee}
\Delta_2    
= \log p \left( \tilde s_{i(u)}(r)| {\bf c} (\tilde{\bf s}_{i(u)}, \hat{\bf d}_{\hat f(i)}, r) ;  \hat{\boldsymbol \phi} \right) \nonumber \\
-\log p \left( s_{i(u)}(r)| {\bf c} ({\bf s}_{i(u)}, \hat{\bf d}_{\hat f(i)}, r) ;  \hat{\boldsymbol \phi} \right) \, ,
\end{align}

With the discussion above, we can update the $u^{th}$ pixel as,  
\begin{equation}
\label{eq:xupdate}
\hat x_u = \argmin_{x_u \in \{0,1\}} \Delta_1 + \Delta_2 \, .
\end{equation}

As shown in Fig. \ref{fig:pseudocode}, we repeat the parameter estimation, dictionary construction, and image restoration for multiple times until convergence,
or a predefined maximum number of iterations is reached due to computing time reason.  
After convergence, we encode the restored image $\hat{\bf x}$ using the JBIG2 lossless option. 
The value of Eq. (\ref{eq:MAP}) is guaranteed to keep decreasing during the optimization procedure. 
We can not guarantee the global optimum due to a lack of convexity, 
but experimental results show that the local optimum we obtained is promising. 


\section{Experimental result}
\label{sec:exp}
In this section, we present all the methods for comparison, and list all the parameter values we have used. 
We conducted experiments with both synthetic noise and real noise to evaluate the performance of our method 
in terms of both image quality 
and compression ratio. 

\subsection{Methods for comparison}
\label{sec:methodlist}
We investigated four cutting-edge methods in our paper.
All these methods follows symbol-dictionary framework in JBIG2 with lossless mode. 

The first two methods encode the observed image (input)
{\bf without restoration}. 
The major difference between these two methods is the way they construct dictionary for encoding:
one method learns the dictionary based on the weighted-XOR dissimilarity measurement (WXOR-Lossless) \cite{P:Combined, M:JBIG2}, 
while
the other method, called CEE-Lossless, learns a dictionary based on the conditional entropy estimation \cite{Yandong:JBIG2CEE}. 

The other two methods encode the {\bf restored image} estimated from the observed image. 
One is the method we proposed in this paper,
called model-based iterative restoration with dictionary learning
(MBIR-DL). 
In our MBIR-DL method, we fixed the matrix $A$ in Eq. (\ref{eq:likelihood_lowp})
as a Gaussian filter with $\sigma_r^2 = 0.2$ throughout all the experiments,  
and applied the JBIG2 lossless mode after the restoration. 

In order to emphasize the benefits from the dictionary used in MBIR-DL,  
we replace the dictionary prior in our MBIR-DL 
with a 
standard Markov Random field (MRF) for binary signals using the $8$-pixel neighborhood system, 
defined in Eq. (\ref{eq:BMRF}), 
\begin{equation}
\label{eq:BMRF}
p(x_k) \propto \exp \left\{ - \sum_{ \{l, k\} \in C} \left| x_k - x_l \right| \right\} \, .
\end{equation}
We call this method MBIR-MRF. 
After its restoration, 
MBIR-MRF encodes the restored image using the same way as MBIR-DL. 
These methods are summarized in Tab. \ref{table:methodlist}. 
\begin{table}
\begin{center}
\begin{tabular}{|l|l|l|}
\hline
Method & Restoration & Encoding Dict.\\
\hline
\\[-1em]
\hline
WXOR-Lossless & No & WXOR \cite{P:Combined, M:JBIG2} \\
CEE-Lossless & No & CEE \cite{Yandong:JBIG2CEE}\\
MBIR-MRF & Yes,  MRF prior& CEE \cite{Yandong:JBIG2CEE} \\
MBIR-DL & Yes,  dictionary prior& CEE \cite{Yandong:JBIG2CEE} \\
\hline
\end{tabular}
\end{center}
\caption{The methods for comparison. 
The first two methods (WXOR-Lossless and CEE-Lossless) encode the input observed image as it is. 
The other two methods encode the restored image estimated from the observed image. 
Our method MBIR-DL restores the observed image with a dictionary prior, while MBIR-MRF uses Markov Random field as prior. 
In regards of encoding, all these methods follow the symbol-dictionary framework in JBIG2 with lossless mode. 
The WXOR-Lossless method encodes image with a dictionary learned based on Weighted-XOR (WXOR) dissimilarity measurement. 
The rest three methods use the same method 
(conditional entropy estimation (CEE) described in \protect\cite{Yandong:JBIG2CEE}) 
to construct
the dictionary for encoding. 
}
\label{table:methodlist}
\end{table}

\subsection{Synthetic noise}

We generate test images with synthetic noise 
so that  
we can evaluate the quality of the restored image 
with a perfectly aligned, noise-free reference image. 
Let ${\bf x}$ denote the reference image (noise free),
and ${\hat{\bf x}} $ denote the restored image estimated from the observed noisy image, 
we count the total number of different pixels between ${\bf x}$ and ${\hat{\bf x}}$ as our quality metric, defined as
\begin{equation}
e= \sum_k \left| \hat x_{k}-  x_k \right| \, ,  
\end{equation}
where $k$ is the pixel index. 
Note that
for a scanned image with inherent real noise,
it is very difficult to obtain a perfectly aligned, noise free reference image
(even the original document pdf is available). 

\subsubsection{Data generation}
We obtain the noise free reference image ${\bf x}$ from the web. 
First, we downloaded pdf files of curriculum vitae of well-known professors. 
\footnote {Due to space limit,
we publish the test data and more detailed experimental results in supplementary materials.} 
Then, we rastered them into binary document images with the the resolution $3240 \times 2550$. 
Together, there are $114$ binary document images containing mainly text. 

In order to synthesize the noise introduced during the imaging procedure, 
we applied a 
Gaussian low-pass filter to each of the test images, which corresponds to $A$ in Eq. (\ref{eq:likelihood_lowp}). 
Note that a similar Gaussian filter is implemented in the firmware in many commercial products, such as Multi-functional printers (MFP).
We followed the same noise model in Eq. (\ref{eq:likelihood_indi}) to generate the scanned image ${\bf y}$. 
Since different value of $\sigma$ lead to different blurry levels and introduce different levels of distortions, 
in our experiment, we applied a $3 \times 3 $ size Gaussian filter with $\sigma^2=0.1$, $0.12$, $0.14$, and $0.16$ to simulate different levels of noise introduced during the imaging process. 
Then we obtained $4$ groups of noisy images with different noisy levels. 

\subsubsection{Compare with compression without restoration}
\begin{figure}[t]
\centering
\hspace{\fill}
\subfigure[Original]{\includegraphics[width=0.235\linewidth]{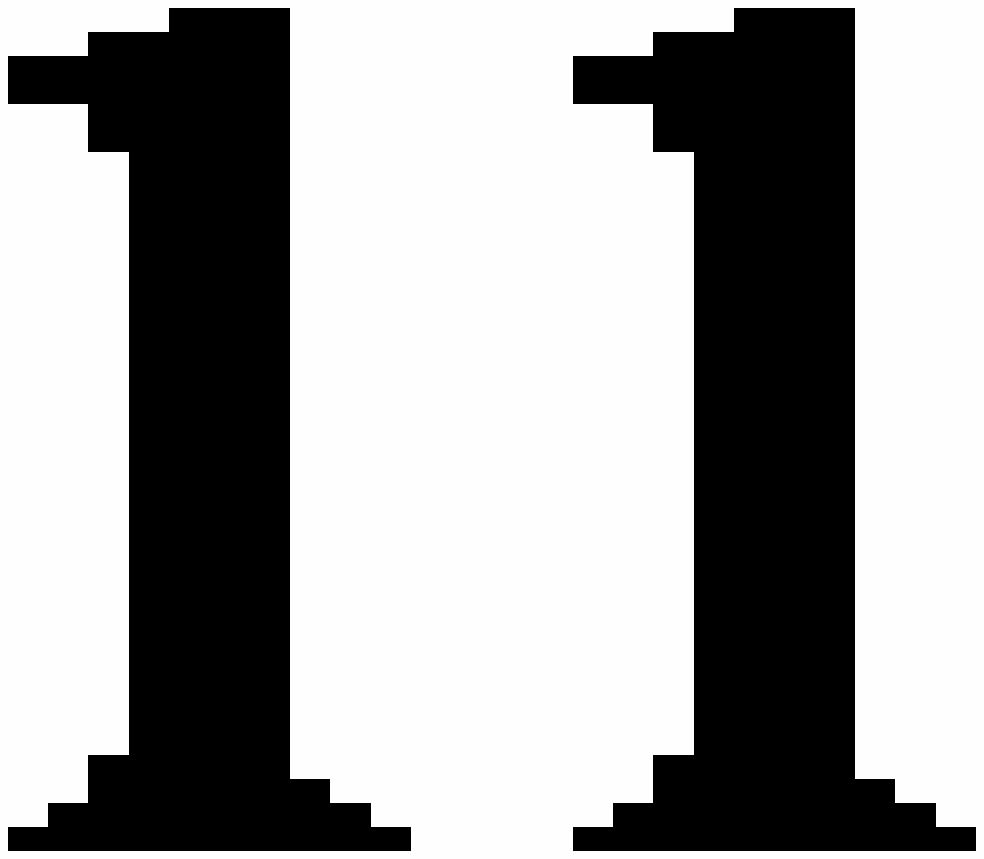}}
\subfigure[Noisy]{\includegraphics[width=0.235\linewidth]{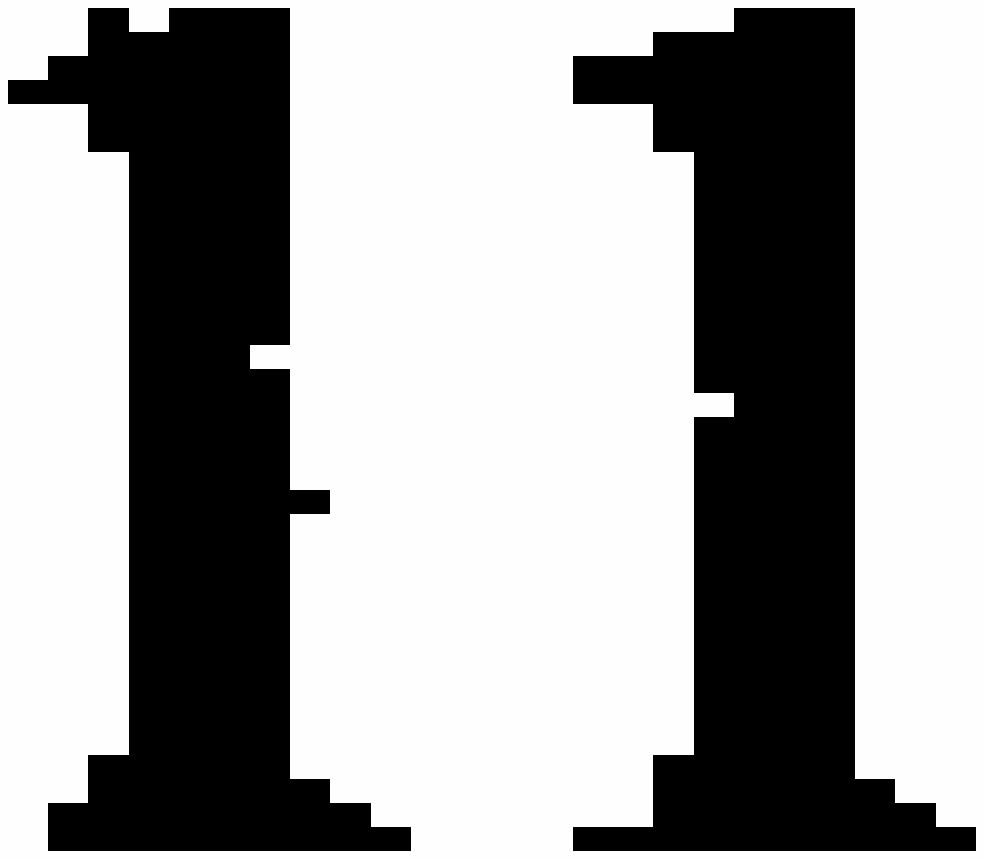}}
\subfigure[MBIR-MRF]{\includegraphics[width=0.235\linewidth]{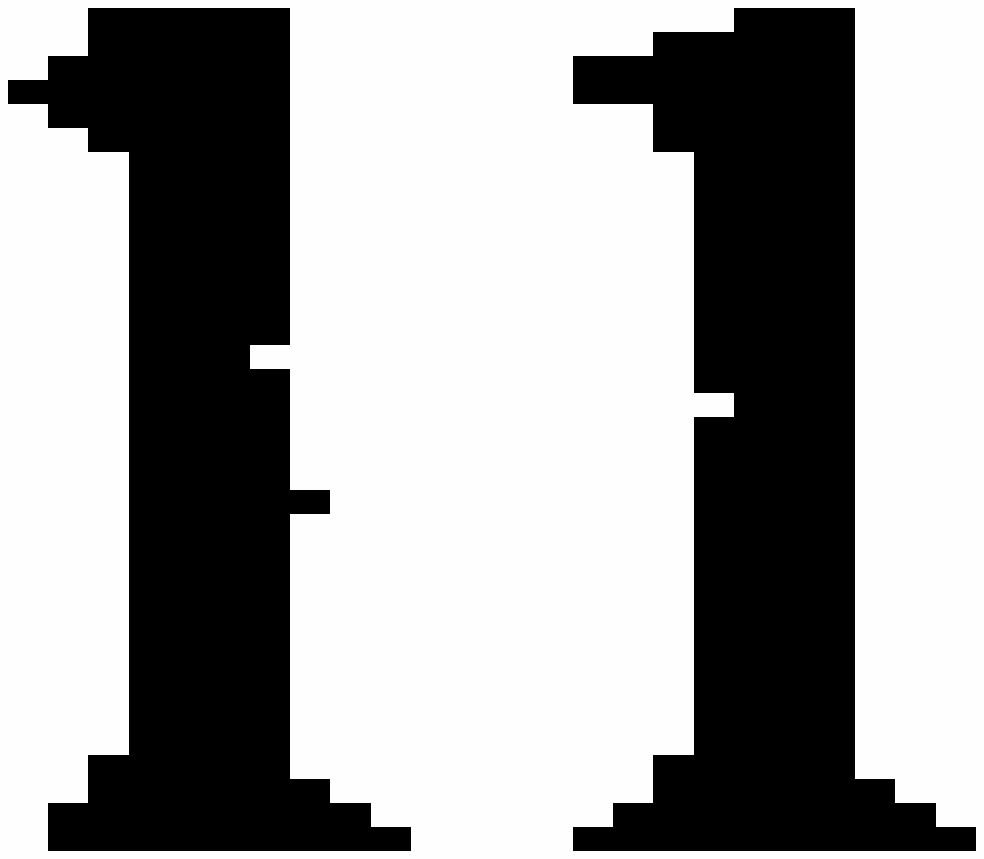}}
\subfigure[MBIR-DL]{\includegraphics[width=0.235\linewidth]{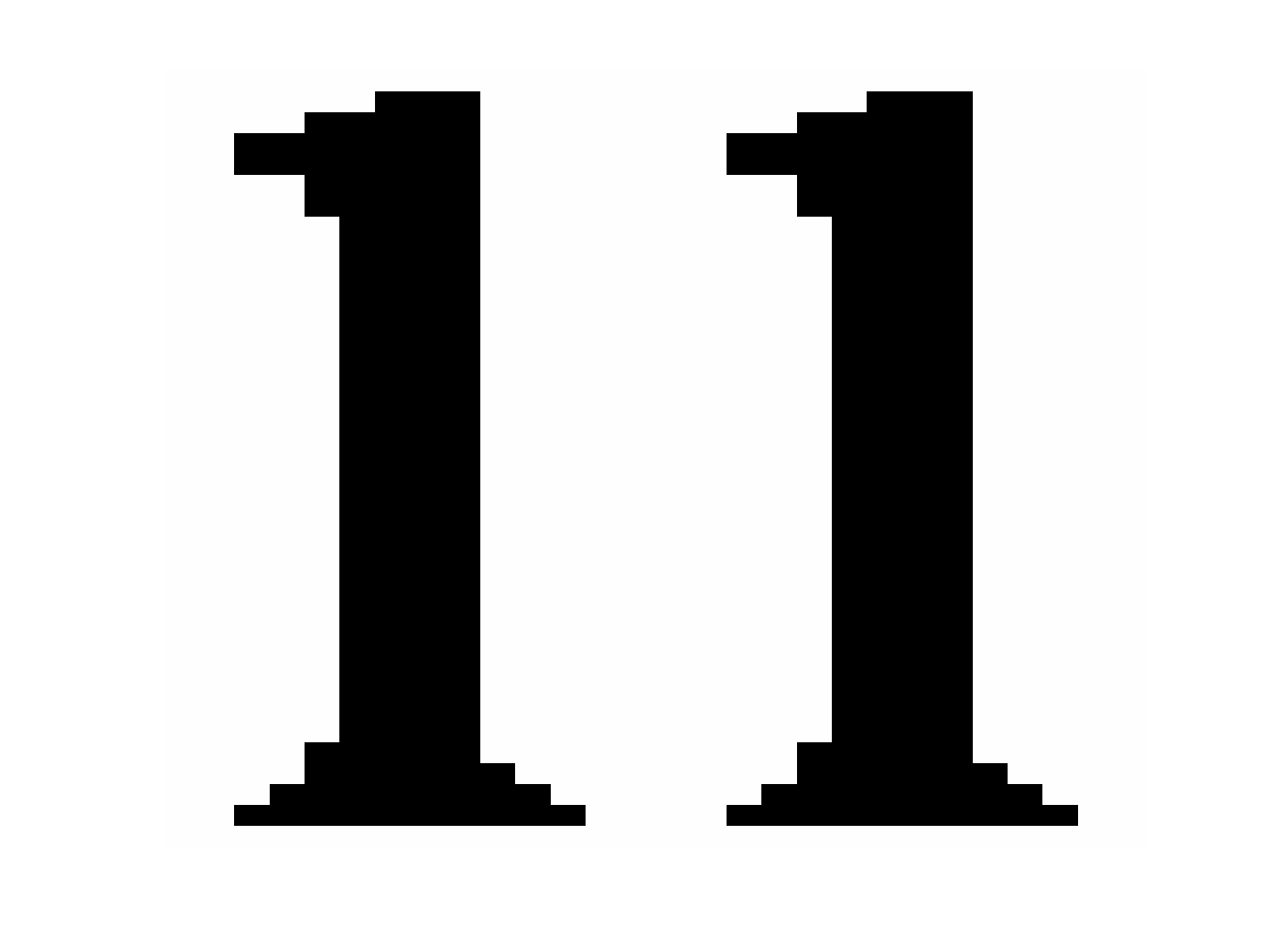}}
\hspace{\fill}
\subfigure[Original]{\includegraphics[width=0.23\linewidth]{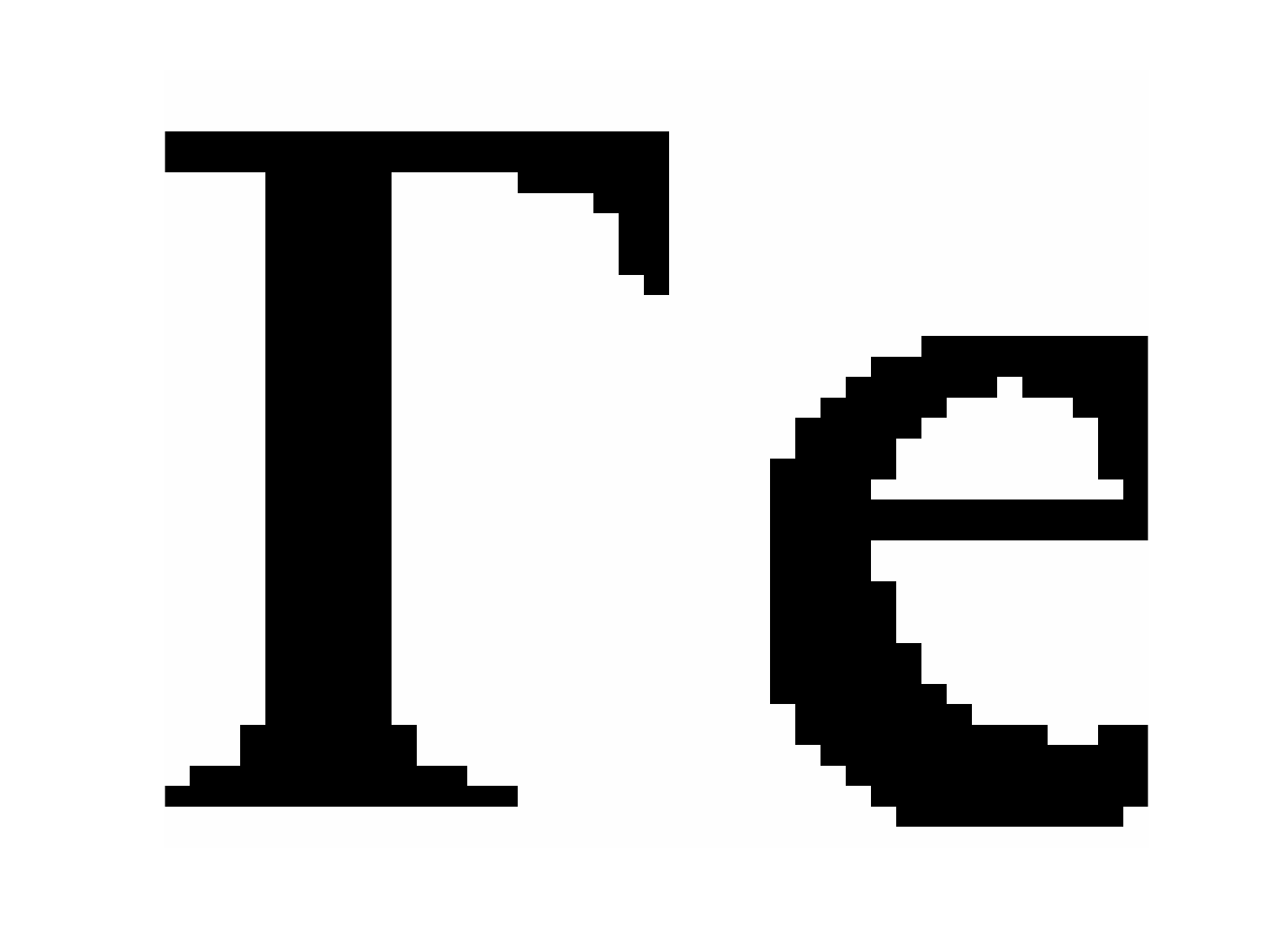}}
\subfigure[Noisy]{\includegraphics[width=0.23\linewidth]{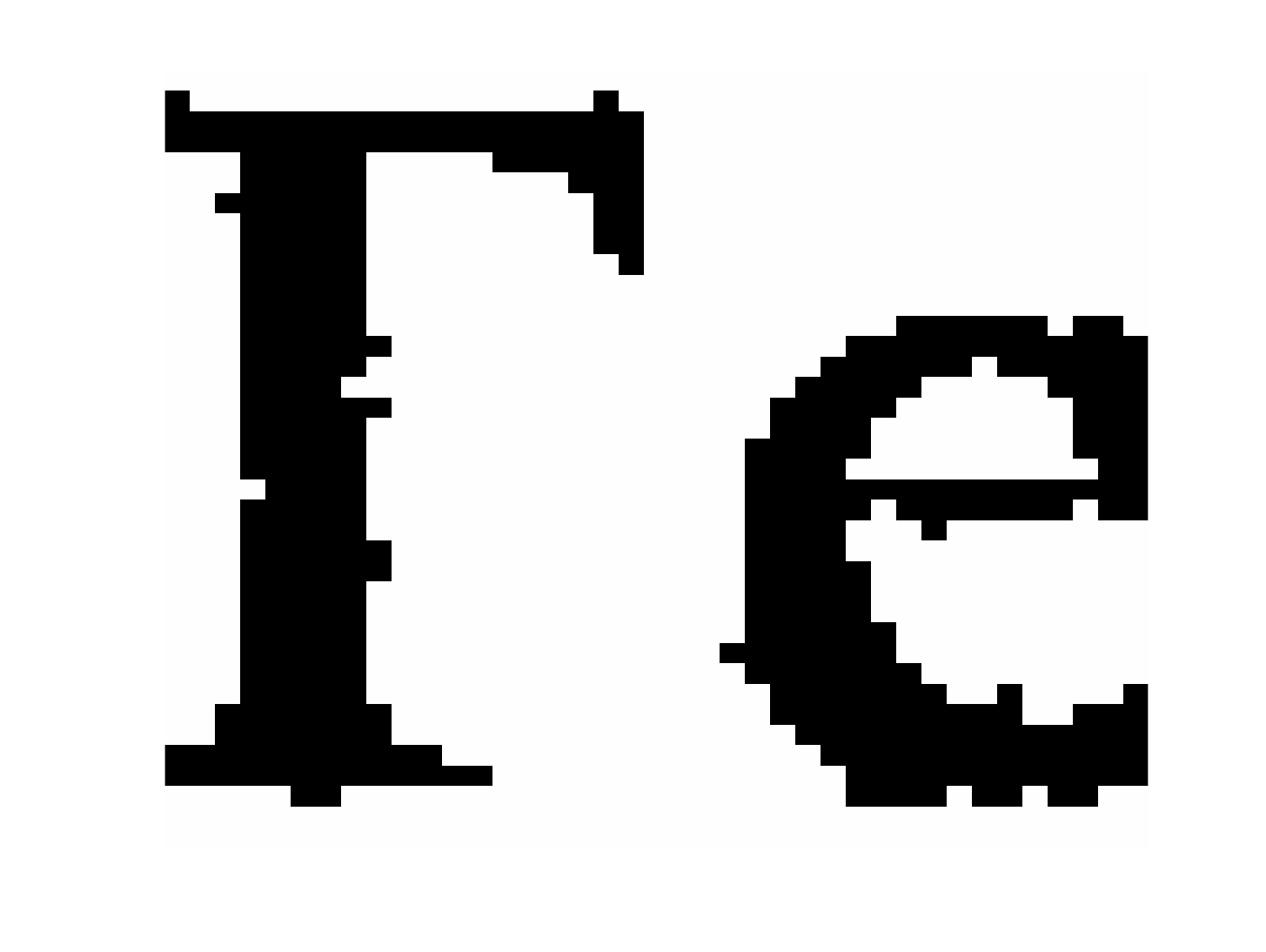}}
\subfigure[MBIR-MRF]{\includegraphics[width=0.23\linewidth]{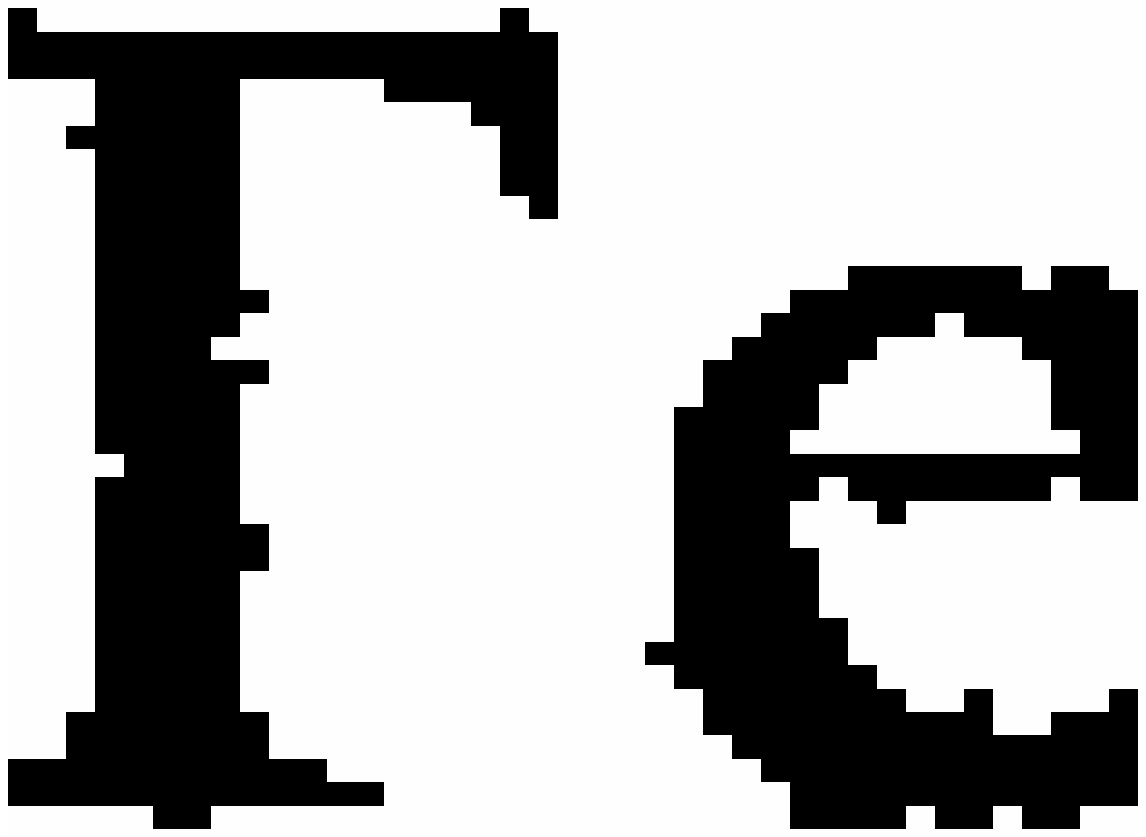}}
\subfigure[MBIR-DL]{\includegraphics[width=0.23\linewidth]{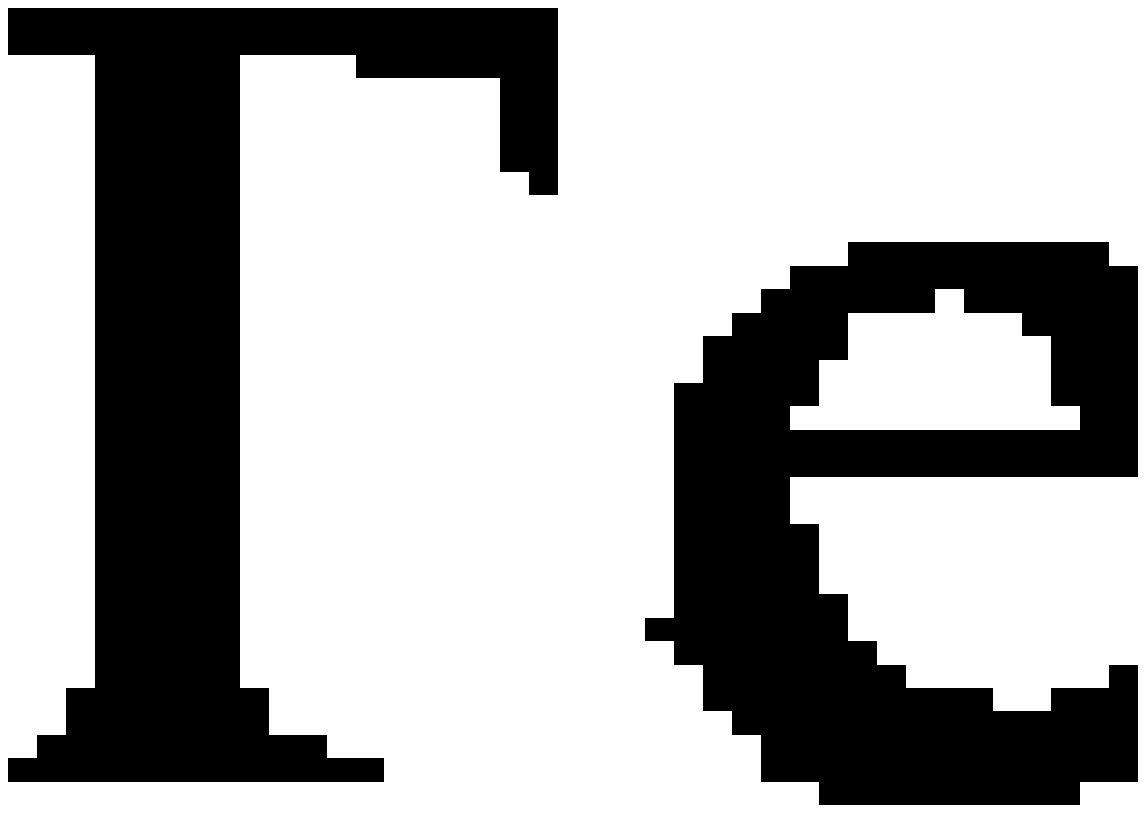}}
\caption{Visualization of the restoration results obtained by using MBIR-DL and MBIR-MRF. We re-list the example of letter ``l'' in Fig. \ref{fig:flowchart} (a) here
for the convenience of the comparison between MBIR-MRF and MBIR-DL.}
\label{fig:recon} 
\end{figure}

\begin{figure}[t]
\centering
\hspace*{\fill}
\subfigure[Number of error pixels $e$]{\includegraphics[width=0.48\linewidth]{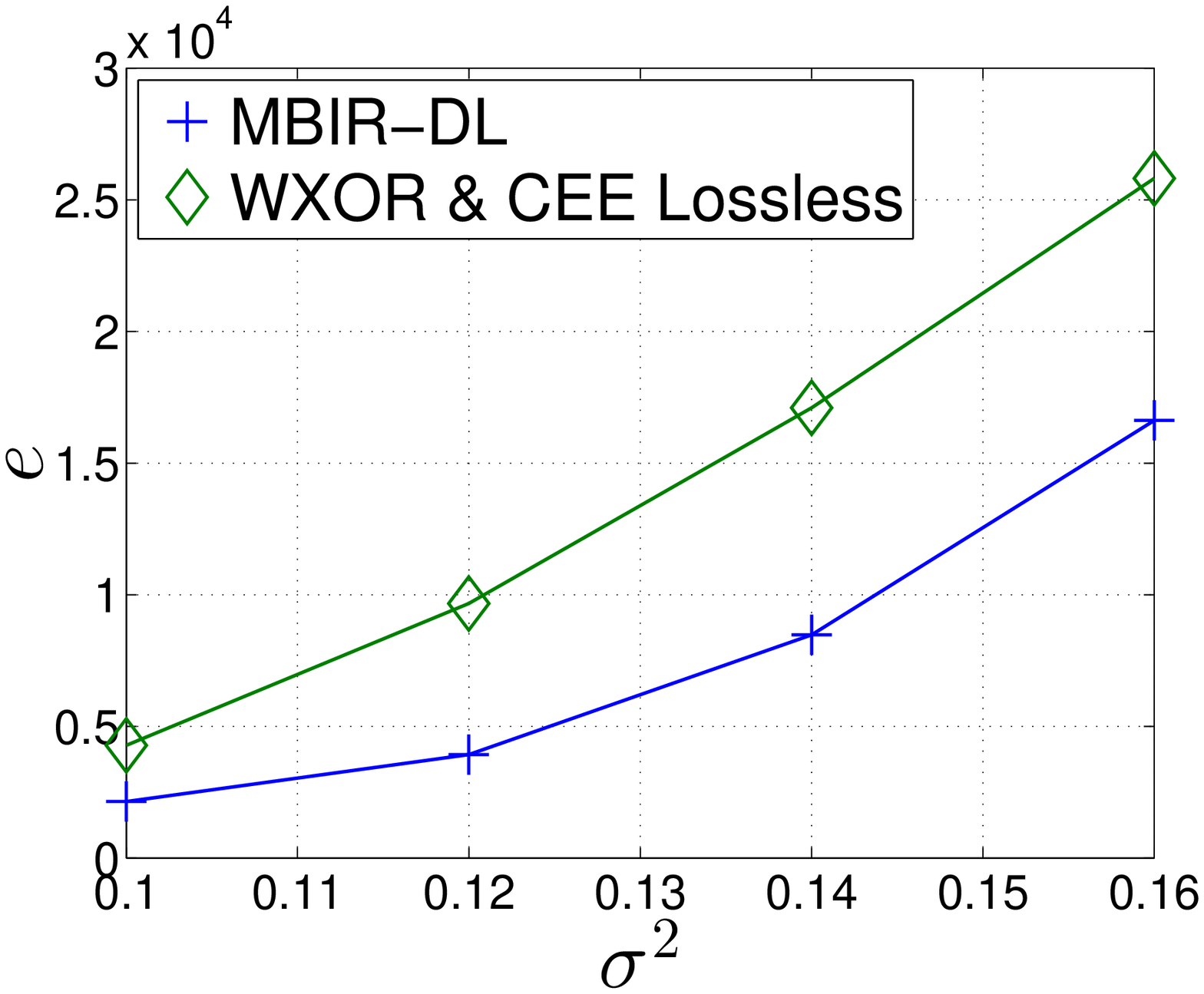}}
\hspace*{\fill}
\subfigure[File size]{\includegraphics[width=0.48\linewidth]{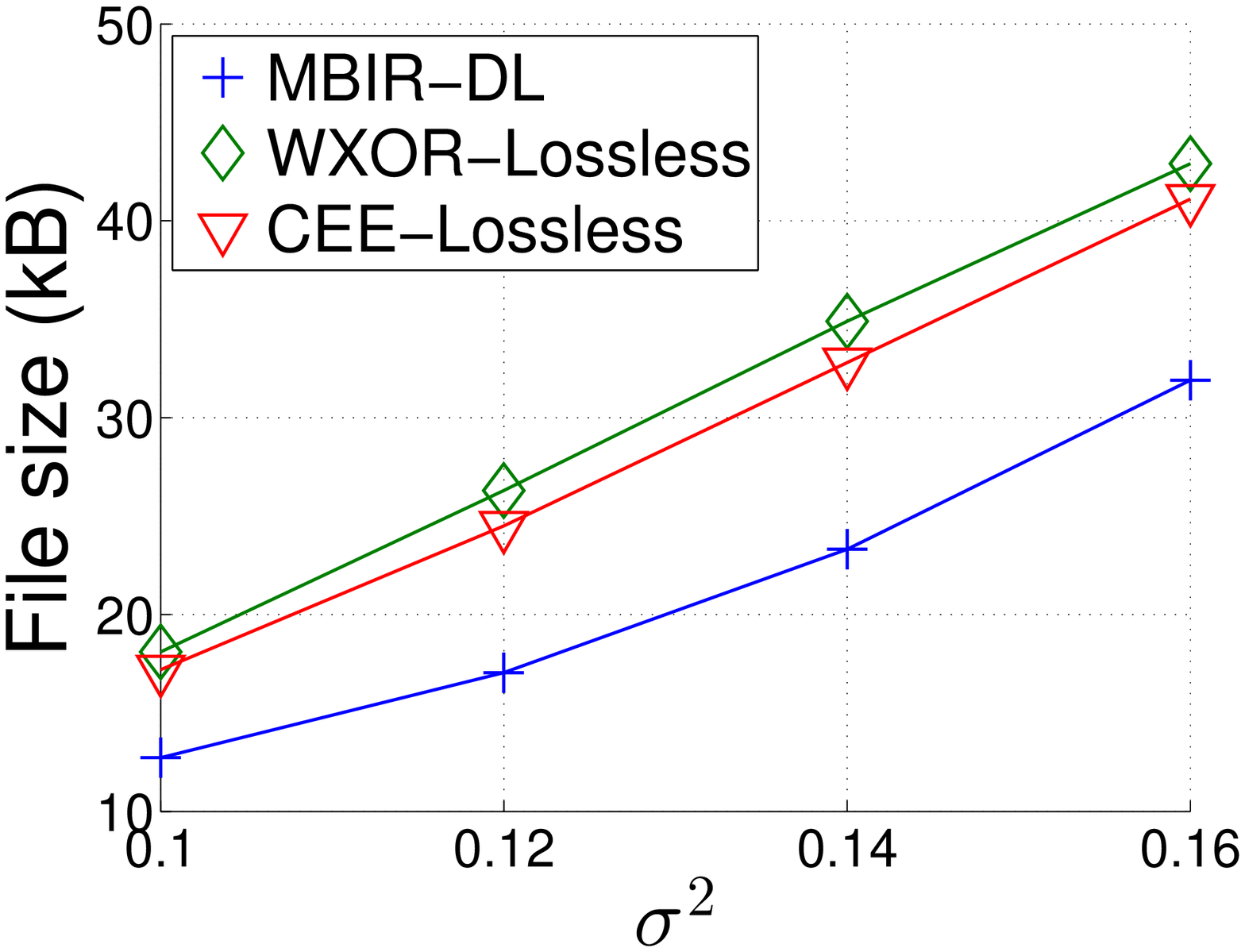}}
\hspace*{\fill}
\caption{Comparison between our MBIR-DL and WXOR-Lossless, CEE-Lossless. 
Neither WXOR-Lossless nor CEE-Lossless change the pixel value of the input image and they have the same quality.
Our MBIR-DL improves image quality and reduces the  file size of the bitstream. 
Note that more noise (larger $\sigma^2$) generally increases file size.}
\label{fig:criq} 
\end{figure}

We compare 
our method 
with WXOR-Lossless in \cite{P:Combined, M:JBIG2}
and CEE-Lossless in \cite{Yandong:JBIG2CEE}. 
Both WXOR-Lossless and CEE-Lossless encode the observed image directly
with the JBIG2 lossless mode.
The quality of 
their compressed image  
is exactly the same as that of the observed image. 
On the contrary, our MBIR-DL method (parameter fixed)
consistently improves the image quality
for the test images with different noise levels, as shown in Fig. \ref{fig:criq}. 

Moreover, 
our MBIR-DL method also consistently outperforms CEE-Lossless and WXOR-Lossless in terms of image compression ratio. 
This is because MBIR-DL restores the observed images and recovers the pattern repentance. 
Note that the CEE-Lossless method produces smaller file size compared with the file size with the WXOR-Lossless, 
because the dictionary learned in the conditional entropy space better represents the binary image. 

\subsubsection{MBIR-DL v.s. MBIR-MRF}

In order to demonstrate the benefit from the dictionary used in the restoration of MBIR-DL, 
we compare MBIR-DL with MBIR-MRF. 
As described in \ref{sec:methodlist}, the only difference between the two methods is that MBIR-MRF uses Markov random field (MRF) as prior, while MBIR-DL uses the dictionary as prior. 

\begin{figure}[t]
\centering
\hspace*{\fill}
\subfigure[Error count $e$]{\includegraphics[width=0.48\linewidth]{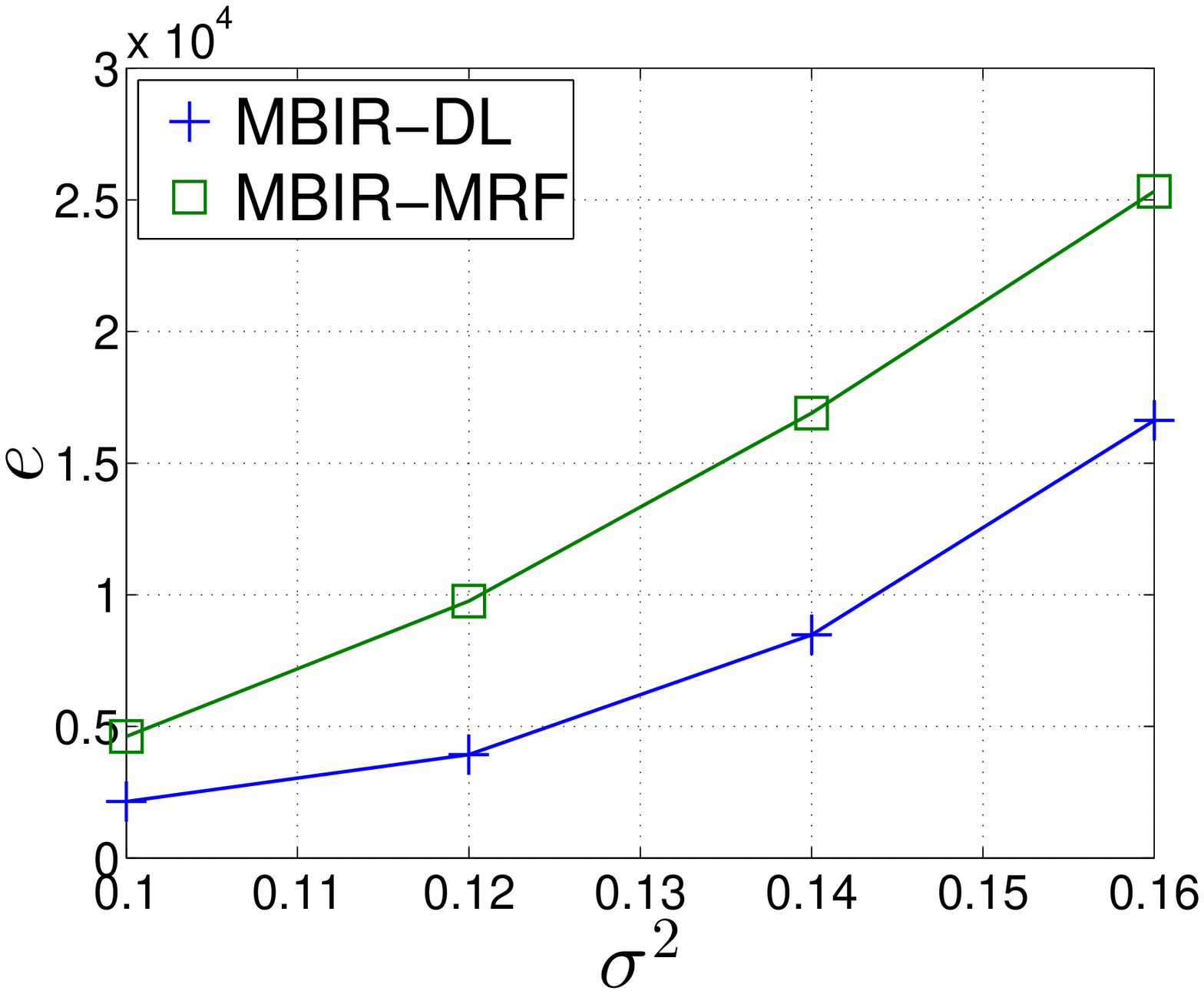}}
\hspace*{\fill}
\subfigure[File size]{\includegraphics[width=0.48\linewidth]{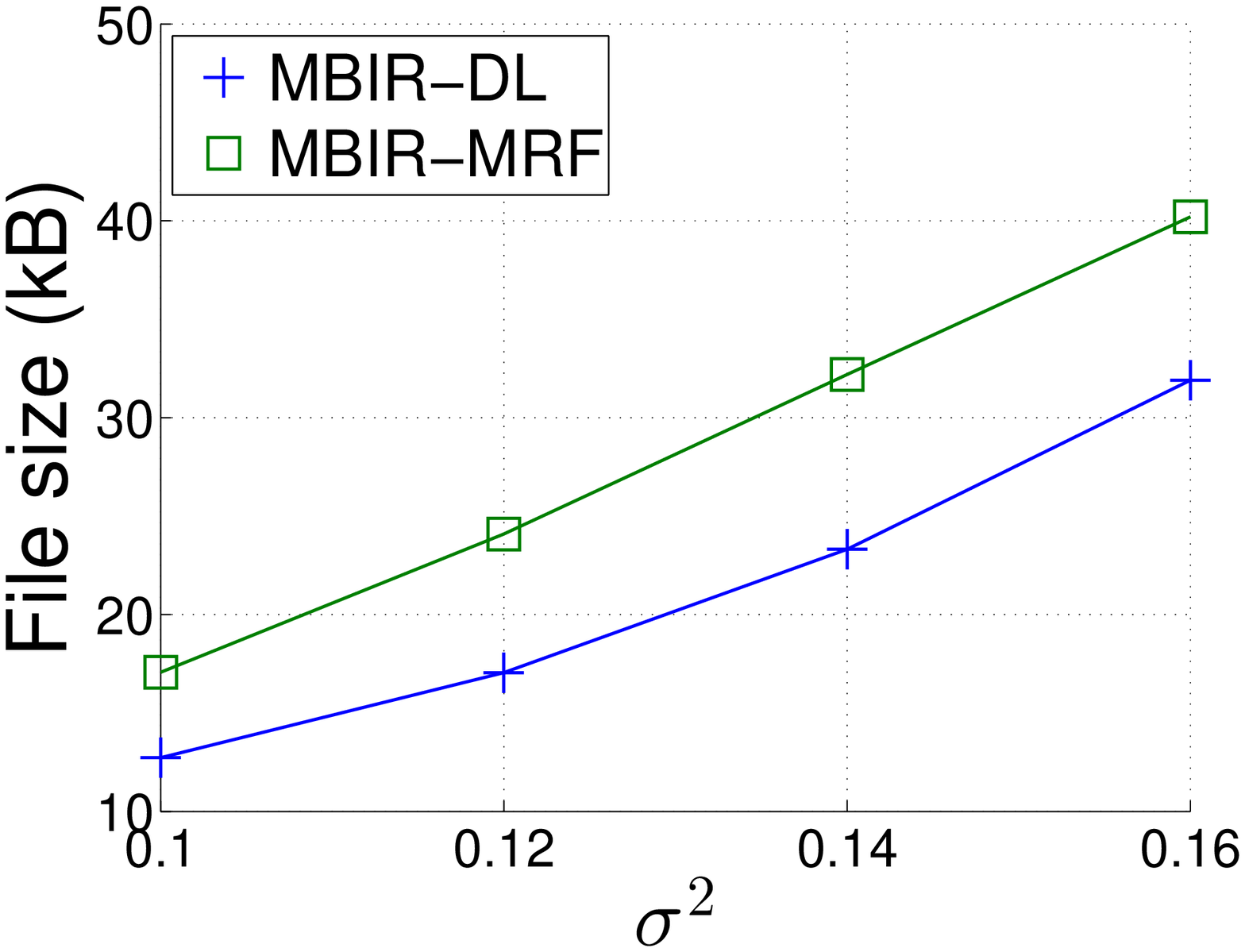}}
\hspace*{\fill}
\caption{The MBIR-DL method outperforms MBIR-MRF
in terms of both image quality and compression ratio. }
\label{fig:criq_recon}
\end{figure}

As shown in Fig. \ref{fig:criq_recon},
our 
MBIR-DL methods outperforms MBIR-MRF in terms of both restoration quality and compression ratio. 
In Fig. \ref{fig:recon}, we visualize the restoration results comparison by zooming in the test images.  
Note that the subfigure (d) is a very typical case that
our MBIR-DL can recover a very sharp left-corner of the left letter ``l'' through the usage of the non-local information. 
However, without non-local information usage, MBIR-MRF does not have the ability to recover this type of fine details with only one pixel wide. 
Also, the subfigures in the last row demonstrate that our MBIR-DL can recover images from severe distortion, though still not perfect.  
 
\subsection{Real noise}


In order to evaluate the performance of our MBIR-DL method in real application scenarios, 
we scanned $41$ binary document images. 
The noise is from the imaging device and more complicated than the synthetic noise. 
All of our test images in this subsection were scanned at 300 dpi, and have size $3275 \times 2525$ pixels.  
These test images contain mainly text, but some of them also contain line art, tables, and generic graphical elements,
but no halftones. 
The text in these test images has various typefaces and font sizes.  

\begin{figure}[t]
\centering
\hspace*{\fill}
\subfigure[Original image]{\includegraphics[width=0.31\linewidth]{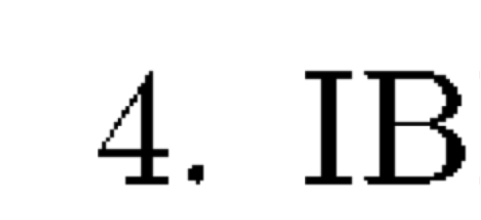}}
\hspace*{\fill}
\subfigure[Scanned image]{\includegraphics[width=0.31\linewidth]{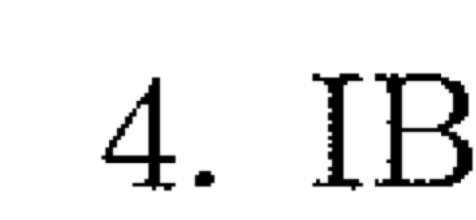}}
\hspace*{\fill}
\subfigure[MBIR-DL]{\includegraphics[width=0.31\linewidth]{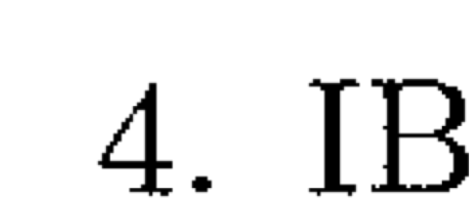}}
\hspace*{\fill}
\caption{Visualization of the restored image obtained by using MBIR-DL}
\label{fig:realQuality} 
\end{figure}

As shown in Tab. \ref{tab:CR}, MBIR-DL achieves the highest compression ratio among all the competitors. 
Since there is no reference image, we evaluate the image quality with non-reference metrics. 
Using the non-reference metric specifically define for binary document images in 
\cite{Lu:Metric}, 
we demonstrate that the visual quality of our restored image has been improved by $5.1\%$. 
We zoomed in to sample areas in the test image for better visualization, 
as shown in Fig. \ref{fig:realQuality}. 
Moreover, we verified the compressed images using both tesseract-OCR and human visual check
for each of the symbols in the image. 
No substitution error was found in the MBIR-DL compressed image.  

\begin{table}
\begin{center}
\begin{tabular}{|l|c|c|}
\hline
\rule[-1ex]{0pt}{3.5ex} Method & File size (KB) & Compression ratio \\
\hline & & \\[-1em]\hline
\rule[-1ex]{0pt}{3.5ex} Lossless-TIFF                         		          & 53.7 KB  &  19.37 \\
\hline
\rule[-1ex]{0pt}{3.5ex} XOR-Lossless                                                    & 35.4 KB   &  29.36 \\
\hline
\rule[-1ex]{0pt}{3.5ex} CEE-Lossless  & 27.8 KB  &  37.40 \\
\hline
\rule[-1ex]{0pt}{3.5ex} MBIR-MRF  & 27.3 KB  &  38.08 \\
\hline
\rule[-1ex]{0pt}{3.5ex} MBIR-DL  & 21.5 KB  &  48.01\\
\hline
\end{tabular}
\end{center}
\caption{Bitstream file size obtained by using different methods to 
the scanned test images with real noise}\label{tab:CR}
\end{table}

\section{Conclusion}
\label{sec:con}
We propose a model-based iterative restoration with dictionary learning method 
to solve a joint optimization regards of image quality and compression ratio. 
By reducing the inevitable noise introduced during the imaging process, including printing, scanning and quantization,
our method simultaneously 
improves the image quality
and compression ratio substantially,
compared directly encoding the observed image input). 
For the test images with synthetic distortion, 
our method 
reduced the number of flipped pixels by $48.2\%$, 
improves the compression ratio by $36.36\%$
as compared to the cutting-edge methods. 
For the test images with real distortion, 
our method outperforms the cutting-edge compression method 
by $28.27\%$ in terms of the compression ratio. 



{\small
\bibliographystyle{ieee}
\bibliography{jbig3}
}

\end{document}